\documentclass[10pt,twocolumn,letterpaper]{article}

\usepackage{cvpr}              

\usepackage[dvipsnames]{xcolor}

\usepackage{amsmath,mathcomp}

\newcommand{\norm}[1]{\left\lVert#1\right\rVert}
\newcommand{\best}[1]{\textbf{#1}}

\definecolor{cvprblue}{rgb}{0.21,0.49,0.74}
\usepackage[pagebackref,breaklinks,colorlinks,citecolor=cvprblue]{hyperref}

\def\paperID{12411} 
\def\confName{CVPR}
\def\confYear{2024}

\title{3D View Optimization for Improving Image Aesthetics}

\author{Taichi Uchida\\
{\tt\small taisws5863@gmail.com}
\and
Yoshihiro Kanamori\\
{\tt\small kanamori@cs.tsukuba.ac.jp}
\and
Yuki Endo\\
{\tt\small endo@cs.tsukuba.ac.jp}
\and
University of Tsukuba
}

\begin{document}
\maketitle

\begin{abstract}
Achieving aesthetically pleasing photography necessitates attention to multiple factors, including composition and capture conditions, which pose challenges to novices. Prior research has explored the enhancement of photo aesthetics post-capture through 2D manipulation techniques; however, these approaches offer limited search space for aesthetics. We introduce a pioneering method that employs 3D operations to simulate the conditions at the moment of capture retrospectively. Our approach extrapolates the input image and then reconstructs the 3D scene from the extrapolated image, followed by an optimization to identify camera parameters and image aspect ratios that yield the best 3D view with enhanced aesthetics. Comparative qualitative and quantitative assessments reveal that our method surpasses traditional 2D editing techniques with superior aesthetics.
\end{abstract}

\section{Introduction}
\label{sec:intro}

For visually appealing photography, it is imperative to deliberate on multiple elements, including composition and the conditions under which the photograph is captured. Post-capture, the aesthetic appeal of photographs can be augmented through the application of photo editing software, such as Adobe Photoshop. While professional photographers possess the acumen to factor in elements that enhance the aesthetic value of images, this remains a challenging endeavor for amateur photographers. In light of this, several methodologies have been introduced to automate the process of image composition editing~\cite{10.1145/3123266.3123274, li2018a2, wei-cvpr2018, li2020composing, Hong_2021_CVPR}, with a focus on improving aesthetic quality. Leveraging these advancements, individuals lacking specialized photography knowledge or skills can readily enhance the visual appeal of their images.
However, current techniques mainly rely on 2D processing methods like cropping and stretching, which offer limited flexibility.

In this paper, we introduce an unprecedented approach that utilizes 3D manipulations to enhance image aesthetics further. 
We first outpaint the input image to widen the search space and reconstruct the 3D scene from the extrapolated image.
We then search for parameters, such as camera translation, rotation, and field-of-view angle, for re-rendering the 3D scene to broaden the possibilities for aesthetic improvement.
Image aesthetics is evaluated using a pre-trained network of aesthetics evaluation model~\cite{wei-cvpr2018}.
During the parameter search, we find that a typical gradient-based optimization approach~\cite{DBLP:journals/corr/KingmaB14} easily falls into local minima. We thus introduce a black-box optimization method in evolutionary computation called {\em Covariance Matrix Adaptation Evolution Strategy} (CMA-ES)~\cite{hansen2001completely} to search for optimal parameters globally.

Our contributions are summarized as follows:
\begin{enumerate}
    \item A pioneering insight to explore camera parameters and image aspect ratios to improve the aesthetics of the input image, and
    \item A simple optimization procedure to search for optimal parameters globally.
\end{enumerate}
Our approach provides a greater degree of freedom in enhancing aesthetics compared to traditional 2D methods. Through both qualitative and quantitative evaluations, we demonstrate the effectiveness of our method in elevating the aesthetic quality of images, showcasing its superiority in visually appealing photography.

\section{Related Work}
\label{sec:RelatedWork}

\subsection{Aesthetics-guided Composition Editing}

Contemporary studies on automatic composition editing for aesthetic enhancement predominantly focus on cropping techniques that trim input images. Such conventional cropping methods are constrained to adjusting the inner regions of an image for editing, particularly when the primary subject is close to the image's boundary, thus limiting the scope for aesthetic improvements. Zhong \etal~\cite{zhong2021aesthetic} overcame this limitation by integrating a pre-cropping extrapolation of the area outside the original frame, thereby expanding the search space for aesthetic optimization.

Another method of changing the composition of an image is through {\em image retargeting}~\cite{DBLP:journals/tog/AvidanS07,10.1145/1882261.1866186}, which resizes an image without distortion while preserving the shape of the main subject. Liu \etal~\cite{DBLP:conf/cae/LiuJW10} proposed a method that automatically performs retargeting while moving the salient subjects in the image to more aesthetically pleasing positions.

While existing approaches to enhancing image aesthetics predominantly employ 2D image processing techniques, our method ventures into an unexplored domain to improve the aesthetics of input images through 3D scene exploration; we optimize camera position, gazing direction, field-of-view angle, and image aspect ratios, all the while preserving the 3D structure of the captured scene.

\subsection{Composition Editing with Depth Information}

Traditional photo-editing tools struggle to incorporate 3D scene information, often failing to preserve spatial relationships between subjects during composition editing. CompZoom~\cite{DBLP:journals/tog/BadkiGKS17} addresses this by allowing manual adjustments to subject size and position through changes in focal length and camera position, utilizing a multi-view camera model and requiring multiple images from different focal lengths and positions of the same scene. Conversely, Liu  \etal~\cite{liu2022zoomshop} introduced a composition-editing tool that leverages depth information from a single image, enabling adjustments in size, position, and perspective of objects while maintaining the scene's 3D structure, albeit necessitating a degree of expertise in photographic composition and familiarity with the tool for effective use. Our approach simplifies this process, offering an automatic composition editing method that manipulates camera parameters within a 3D scene reconstructed from a single image, making it accessible to users without in-depth photography knowledge.

\subsection{Aesthetic View Exploration in 3D Space}
\label{sec:ViewFind3D}

Prior research on aesthetic view search within 3D spaces has predominantly utilized reinforcement learning to ascertain optimal camera positions and orientations in both real and virtual 3D environments. Xie \etal~\cite{xie2023gait} introduced a technique for generating sequences of aesthetically appealing camera postures within a pre-constructed 3D scene, employing the same aesthetics evaluation model as AlZayer \etal~\cite{9636788}. In contrast, our approach focuses on identifying aesthetic views in the 3D scene reconstructed from a single input photograph, offering a novel perspective on enhancing image aesthetics.

\subsection{Aesthetic Evaluation of Images}
\label{sec:AestheticAssessment}

The task of image aesthetics evaluation involves categorizing images into distinct levels of aesthetics or quantifying an image's beauty with an aesthetics score. The advent of deep learning and the availability of extensive datasets for aesthetics assessment have led to deep learning-based methods outperforming traditional, manually crafted approaches in terms of accuracy and efficiency. These methods, noted for their robustness and minimal inference costs, have been effectively integrated into view-search algorithms within 3D spaces, as discussed in Section~\ref{sec:ViewFind3D}. Our approach leverages the aesthetics evaluation model called a {\em View Evaluation Net} (VEN) proposed by Wei \etal~\cite{wei-cvpr2018}, which calculates aesthetics scores for images generated from initial viewpoints and various camera positions. This model, trained on a comprehensive dataset comprising over 1 million photo-to-picture pair comparisons, excels in determining the relative aesthetics among pairs of image crops, providing a solid foundation for our method's aesthetics-based view selection.

\begin{figure*}[t]
\centering
    \includegraphics[width=\textwidth]{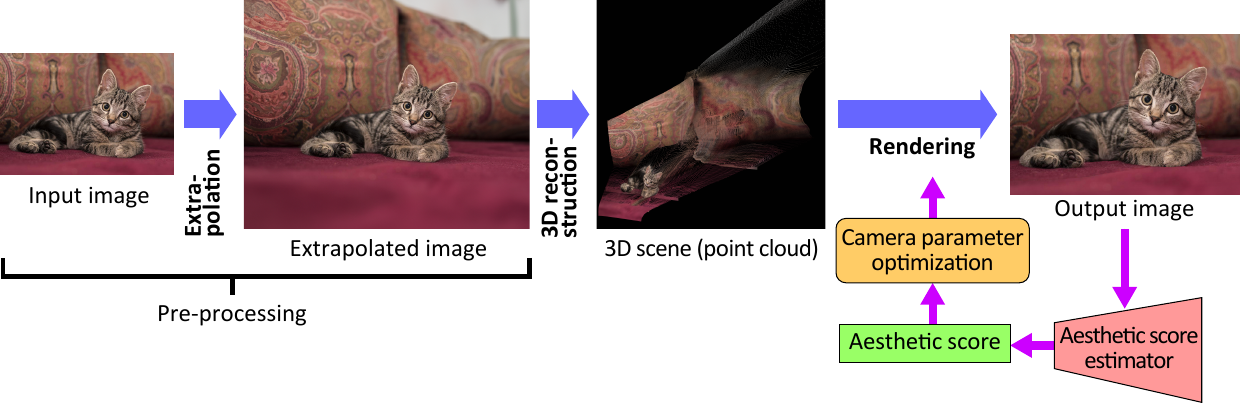}
\caption{Overview of our method. The input image is first extrapolated in the pre-processing and then fed to a 3D reconstruction method~\cite{Shih3DP20} to obtain a point cloud that represents the 3D scene. Our method finds optimal camera parameters that maximize the aesthetic score~\cite{wei-cvpr2018} of the rendered image through an optimization loop.}
\label{fig:Overview}
\end{figure*}

\section{Our Method}
\label{sec:OurMethod}

Figure~\ref{fig:Overview} illustrates the overview of our method.
Our method commences with an image extrapolation of the input image utilizing Adobe Photoshop's generative fill function (Section~\ref{sec:Extrapolation}). Subsequently, a depth map is generated from the extrapolated image via a monocular depth estimation technique~\cite{Ranftl2020}. Direct restoration of the 3D scene from the input image and depth map would reveal regions beyond the original field of view upon altering the viewpoint. We thus feed the extrapolated image and corresponding depth map into a 3D photo generation method~\cite{Shih3DP20} to reconstruct the 3D scene, resulting in a point cloud that supplements the color and depth information for the unseen areas of the input image.
We then search for camera parameters, \ie, translation, rotation, and field-of-view angle, to identify configurations that enhance aesthetic quality. An existing aesthetic evaluation model~\cite{wei-cvpr2018} assesses the aesthetics of images rendered from new viewpoints within this 3D scene. The optimization process seeks to maximize the aesthetic scores by adjusting the camera parameters and also explores variations in the image aspect ratio to broaden the search space. Further details on each step of our method are elaborated in the subsequent sections.

\subsection{Preliminary Image Extrapolation}
\label{sec:Extrapolation}

If we na\"ively reconstruct a 3D scene from the input image, a slight change in camera position or gazing direction will reveal out-of-photographed regions beyond the edges of the input image.
To avoid this, we extrapolate the input image in advance. 
Specifically, the input image is first uniformly resized such that the larger dimension, whether height $h$ or width $w$ of the input image, measures 512 pixels. Subsequently, we expand the image's borders by 256 pixels on all sides (\ie, top, bottom, left, and right) utilizing Adobe Photoshop's generative fill function (without text prompts). This preparatory step not only enhances the search space for aesthetic improvements but also effectively minimizes the visibility of areas beyond the original photographic range.

\subsection{Suppressing the Appearance of Out-of-photographed Regions}
\label{sec:MaskRegularization}

Unfortunately, the abovementioned image extrapolation is not perfect yet; a large change in camera translation or gazing direction may still reveal out-of-photographed regions.

To further suppress the appearance of such regions, our method integrates a regularization term defined with a binary mask rendered under new camera parameters. Let $\mathcal{P}$ be the colored point cloud obtained from the input image, $\mathbf{\Theta}$ be the camera parameters, and $\mathcal{M}$ be the rendering function that outputs a binary mask. The regularization term is defined as follows:
\begin{equation}
    \mathcal{L}_\mathit{mask} = \frac{1}{w \times h} \! \norm{\mathbf{1}_{w \times h} - \mathcal{M}(\mathcal{P}, \mathbf{\Theta})}^2,
    \label{eq:MaskConstraint}
\end{equation}
where $\mathbf{1}_{w \times h}$ is an all-``1'' image of $w \times h$ pixels.

\subsection{Camera Parameter Optimization}

Let $\mathcal{R}$ be a rendering function that outputs a color image from point cloud $\mathcal{P}$ under the camera parameter $\Theta$, and $\mathcal{A}$ be an evaluation function that outputs an aesthetic score from a color image. We update the camera parameter $\Theta$ using the following equation.
\begin{equation}
    \Theta^* = \underset{\mathbf{\Theta}}{\operatorname{argmax}} \,\, \mathcal{A}(\mathcal{R}(\mathcal{P}, \mathbf{\Theta})) - \lambda_\mathit{mask} \, \mathcal{L}_\mathit{mask},
    \label{eq:Optimization}
\end{equation}
where $\lambda_\mathit{mask}$ is the weight of $\mathcal{L}_\mathit{mask}$ and we set $\lambda_\mathit{mask}=10$.

Initially, we used the differentiable renderer of PyTorch3D~\cite{10.1145/3415263.3419160} as the rendering function $\mathcal{R}$ and the Adam optimizer~\cite{DBLP:journals/corr/KingmaB14} to optimize Equation~\eqref{eq:Optimization}. However, our preliminary experiment showed that this optimization easily falls into local minima.
Therefore, we adopt a black-box optimization algorithm called CMA-ES~\cite{hansen2001completely} for a more global search. Note that this algorithm does not require gradient computation and thus differentiable rendering is not required.

The camera parameters to be optimized in this study are the camera translation vector $\mathbf{t} = (t_x, t_y, t_z)$, camera rotation angle $\mathbf{\theta}=(\theta_\mathit{roll}, \theta_\mathit{pitch}, \theta_\mathit{yaw})$, and vertical field-of-view angle $\theta_\mathit{fovy}$. Furthermore, to change the aspect ratio of the output image, we introduce coefficients $s_w, s_h$ to control the output image size $(s_w w) \times (s_h h)$ and optimize these coefficients. A comparison with and without optimization of these coefficients is presented in the next section.

\section{Experiments}
\label{sec:Experiments}

\subsection{Experimental settings}

We implemented our method using Python, PyTorch, PyTorch3D, and Optuna~\cite{optuna_2019}. Experiments were conducted using the test images of the following three datasets: FCDB~\cite{chen-wacv2017}, FLMS~\cite{10.1145/2647868.2654979}, and GAICD~\cite{8953733}. These three datasets contain 309, 500, and 500 test images, respectively. Image extrapolation was performed on all of the test images in advance. The optimization took approximately 5 minutes per image on an Intel Core i7-5960X CPU.

We adopted CMA-ES as the optimization algorithm because it gave the best results among the algorithms implemented in Optuna.
The number of optimization steps was set to 2,000, and optimization was terminated if the evaluated value did not decrease by 0.001 or more in the last 500 steps.

The search ranges of parameters were set as follows. The camera translation vector was set as $t_x,t_y \in [-0.1,0.1]$ and $t_z \in [-0.5, 0.5]$, the camera rotation angles were $\theta_\mathit{roll},\theta_\mathit{pitch},\theta_\mathit{yaw} \in [-10\tcdegree, 10\tcdegree]$, the vertical field-of-view angle was $\theta_\mathit{fovy} \in [-10\tcdegree, 10\tcdegree]$. The coefficients $s_w, s_h$ were initialized as 1, and the search range was set to $[0.1, 2]$.

\begin{table}[t]
 \caption{Ablation study of our method. We compare three variants, \ie, optimized using (i) Adam~\cite{DBLP:journals/corr/KingmaB14}, (ii) CMA-ES~\cite{hansen2001completely}, and (iii) CMA-ES with image scaling (\ie, image aspect ratio optimization) with three image datasets. Boldface indicates best values.}
 \label{tab:AblationScore}
 \centering
 \footnotesize
  \begin{tabular}{l||ccc|c}
   \hline
   Method & FCDB & FLMS & GAICD & Total \\
   \hline
   Input (before opt.) & 0.84 & 1.18 & 0.77 & 0.94 \\
   Adam & 1.30 & 1.64 & 1.22 & 1.40 \\
   CMA-ES w/o scaling & 3.22 & 3.21 & 2.81 & 3.06 \\
   CMA-ES w/ scaling & \best{3.55} & \best{3.50} & \best{3.11} & \best{3.36} \\
   \hline
  \end{tabular}
\end{table}

\begin{table}[t]
 \caption{Quantitative comparison with existing 2D cropping methods. We calculated aesthetic scores using VEN~\cite{wei-cvpr2018}, SAMPNet~\cite{zhang2021image}, and TANet~\cite{DBLP:conf/ijcai/HeZXJM22} as evaluation metrics.}
 \label{tab:ComparisonScore}
 \centering
  \resizebox{\columnwidth}{!}{%
  \begin{tabular}{l||cccc|cccc|cccc}
   \hline
   & \multicolumn{4}{|c}{VEN scores $\uparrow$}
   & \multicolumn{4}{|c}{SAMPNet scores $\uparrow$}
   & \multicolumn{4}{|c}{TANet scores $\uparrow$} \\ \cline{2-13}
   Method & FCDB & FLMS & GAICD & Total & FCDB & FLMS & GAICD & Total & FCDB & FLMS & GAICD & Total \\
   \hline
   Input & 0.84 & 1.18 & 0.77 & 0.94 & 2.87 & 3.14 & 3.31 & 3.14 & 0.49 & 0.50 & 0.54 & 0.51 \\
   VPN~\cite{wei-cvpr2018} & 0.60 & 0.89 & 0.55 & 0.69 & 2.84 & 3.11 & 3.24 & 3.10 & 0.48 & 0.49 & 0.52 & 0.50 \\
   VEN~\cite{wei-cvpr2018} & 0.86 & 1.12 & 0.82 & 0.94 & 2.86 & 3.11 & 3.27 & 3.11 & 0.48 & 0.49 & 0.52 & 0.50 \\
   CGS~\cite{li2020composing} & 0.23 & 0.46 & 0.11 & 0.27 & 2.80 & 3.10 & 3.24 & 3.08 & 0.48 & 0.48 & 0.52 & 0.49 \\
   CAC~\cite{Hong_2021_CVPR} & 0.30 & 0.60 & 0.22 & 0.38 & 2.81 & 3.10 & 3.24 & 3.08 & 0.48 & 0.49 & 0.54 & 0.50 \\
   UNIC~\cite{Liu_2023_ICCV} & 0.64 & 0.94 & 0.60 & 0.74 & 2.85 & 3.15 & 3.31 & 3.14 & 0.49 & 0.50 & 0.54 & 0.51 \\
   Ours & \best{3.55} & \best{3.50} & \best{3.11} & \best{3.36} & \best{2.96} & \best{3.24}
   & \best{3.35} & \best{3.21} & \best{0.51} & \best{0.52} & \best{0.55} & \best{0.53} \\
   \hline
  \end{tabular}
  }%
\end{table}

\subsection{Ablation Study}

We conducted an ablation study of our method.
Specifically, we compared three variants: optimized using (i) Adam as a gradient-based optimizer, (ii) CMA-ES as a black-box optimizer, and (iii) CMA-ES with scale adjustment (\ie image aspect ratio optimization).
As an evaluation metric, we adopted aesthetic scores estimated using VEN~\cite{wei-cvpr2018}.
While VEN's scores are unbounded, the larger score indicates more aesthetic.
Table~\ref{tab:AblationScore} shows a quantitative comparison of the three variants and the input images.
We can see Adam only slightly improves the scores from the input images, but the margins are quite small.
Contrarily, CMA-ES outperforms Adam by a large margin, and the scores are further improved by accounting for image size adjustment.

\begin{figure*}[t]
\centering
    \includegraphics[width=\textwidth]{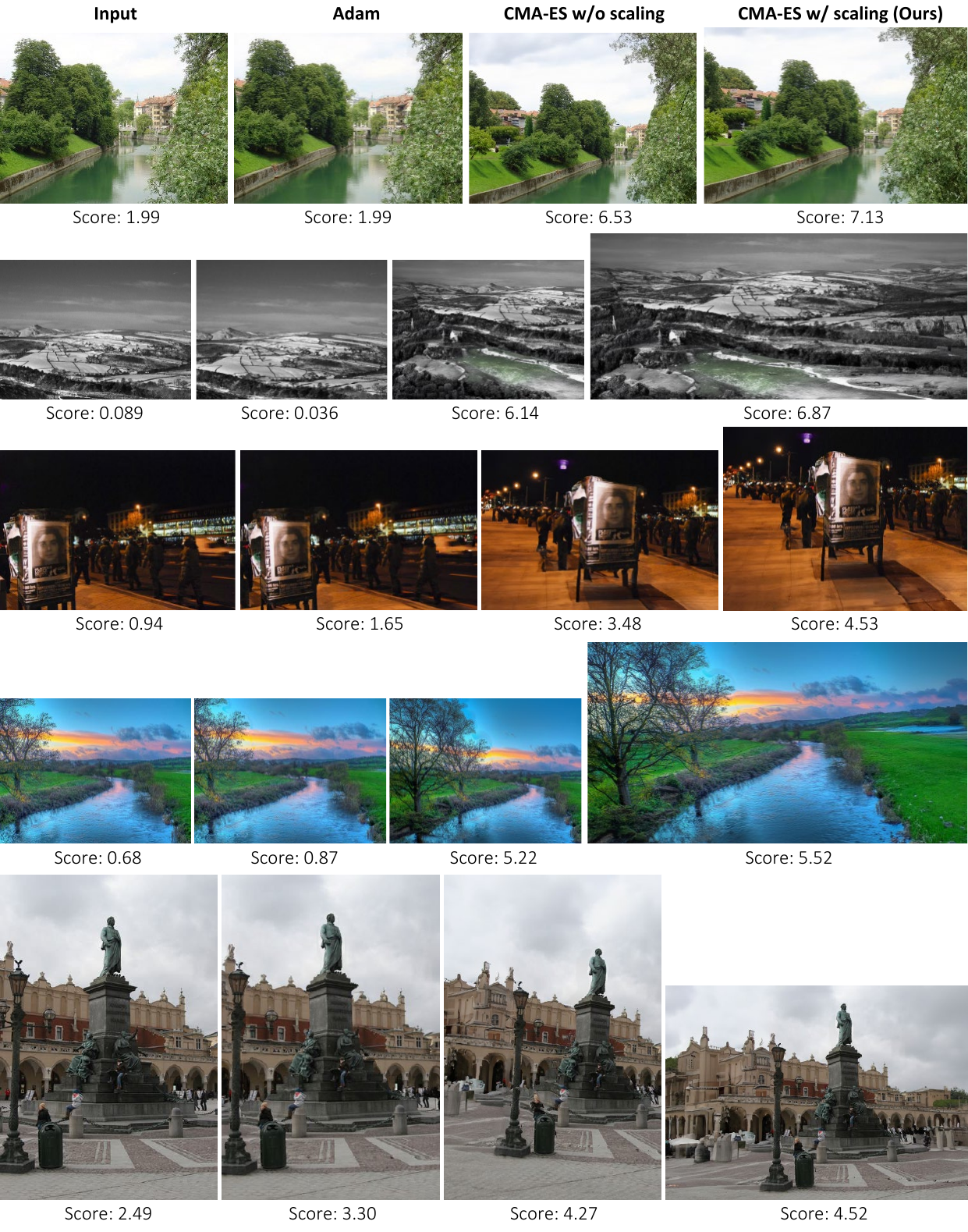}
\caption{Ablation study. We compare three variants; optimized using (i) Adam (2nd column), (ii) CMA-ES (3rd column), and (iii) CMA-ES with image scaling (4th column). The aesthetic scores (the larger, the better) under images are calculated using VEN.}
\label{fig:AblationStudy}
\end{figure*}

This trend can also be confirmed from the qualitative comparison shown in Figure~\ref{fig:AblationStudy}.
Adam improves the scores only slightly but the resultant images look almost the same as the input images, which indicates that Adam fell into local minima.
Using CMA-ES, the scores are drastically improved and the 3D views are largely changed from the input images.
Note that the camera's viewing directions are clearly changed in the first and third rows of Figure~\ref{fig:AblationStudy} while preserving the 3D scene structure, which cannot be accomplished using traditional 2D cropping methods.
The scores are further improved by accounting for the image scaling to adjust image aspect ratios.
Also note that the aspect ratio might change significantly, for example, a vertical image became horizontal; see the bottom example in Figure~\ref{fig:AblationStudy}.
Hereafter we denote CMA-ES with image scaling as our method.

\subsection{Comparison with Existing 2D Cropping Methods}

Next, we compared our method with existing 2D cropping methods.
The compared methods are VPN~\cite{wei-cvpr2018}, VEN~\cite{wei-cvpr2018}, CGS~\cite{li2020composing}, CAC~\cite{Hong_2021_CVPR}, and UNIC~\cite{Liu_2023_ICCV}.
As evaluation metrics, in addition to VEN~\cite{wei-cvpr2018}, we also evaluated aesthetic scores using SAMPNet~\cite{zhang2021image} and TANet~\cite{DBLP:conf/ijcai/HeZXJM22}.

Table~\ref{tab:ComparisonScore} shows the quantitative evaluation result.
Surprisingly, even though our method uses only VEN as an aesthetic score estimator during optimization, our results are consistently better than other methods when evaluated using not only VEN but also SAMPNet and TANet.

\begin{figure*}[t]
\centering
    \includegraphics[width=\textwidth]{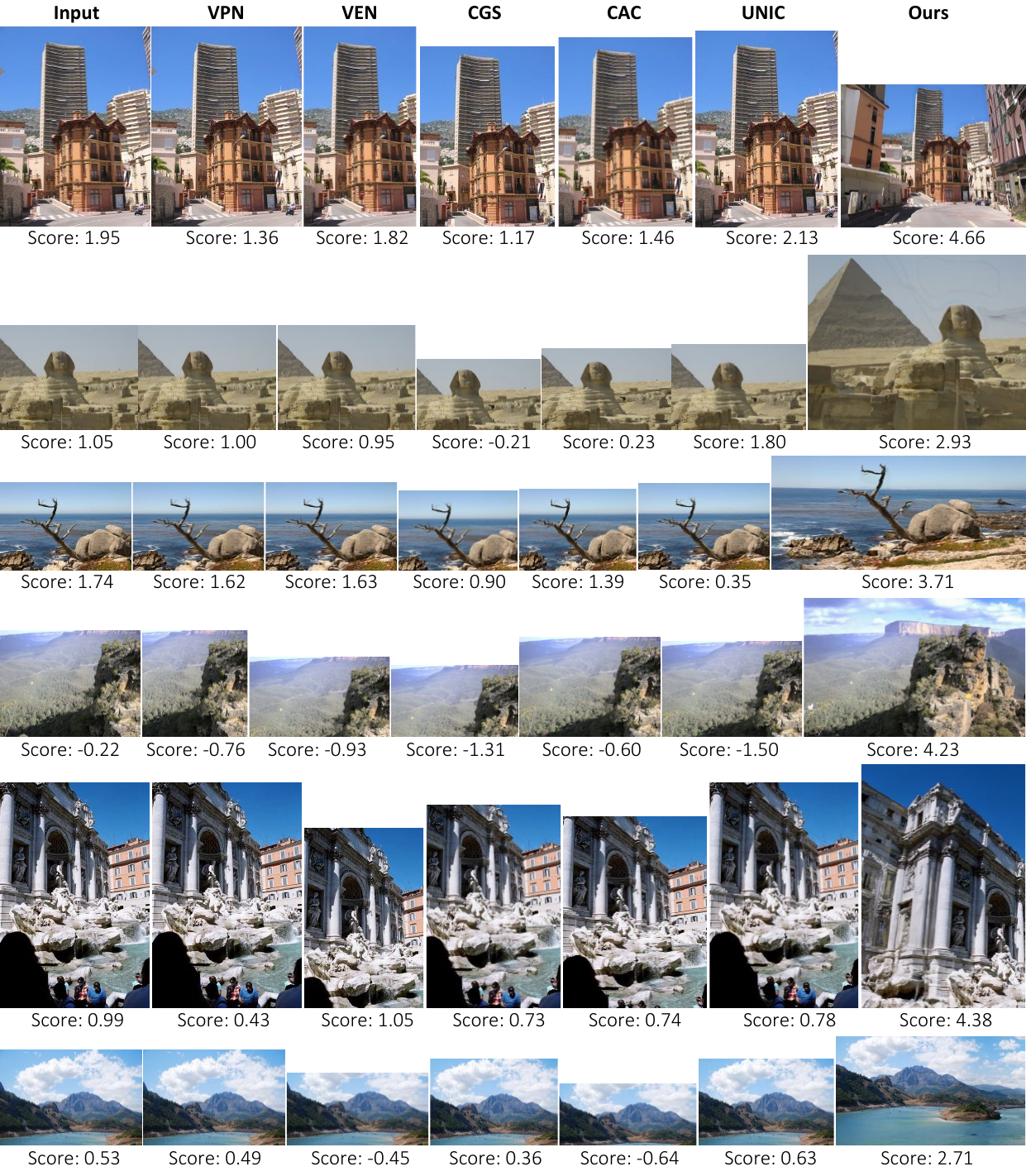}
\caption{Qualitative comparison of our method with existing 2D cropping methods.}
\label{fig:Comparison2}
\end{figure*}

\begin{figure*}[t]
\centering
    \includegraphics[width=\textwidth]{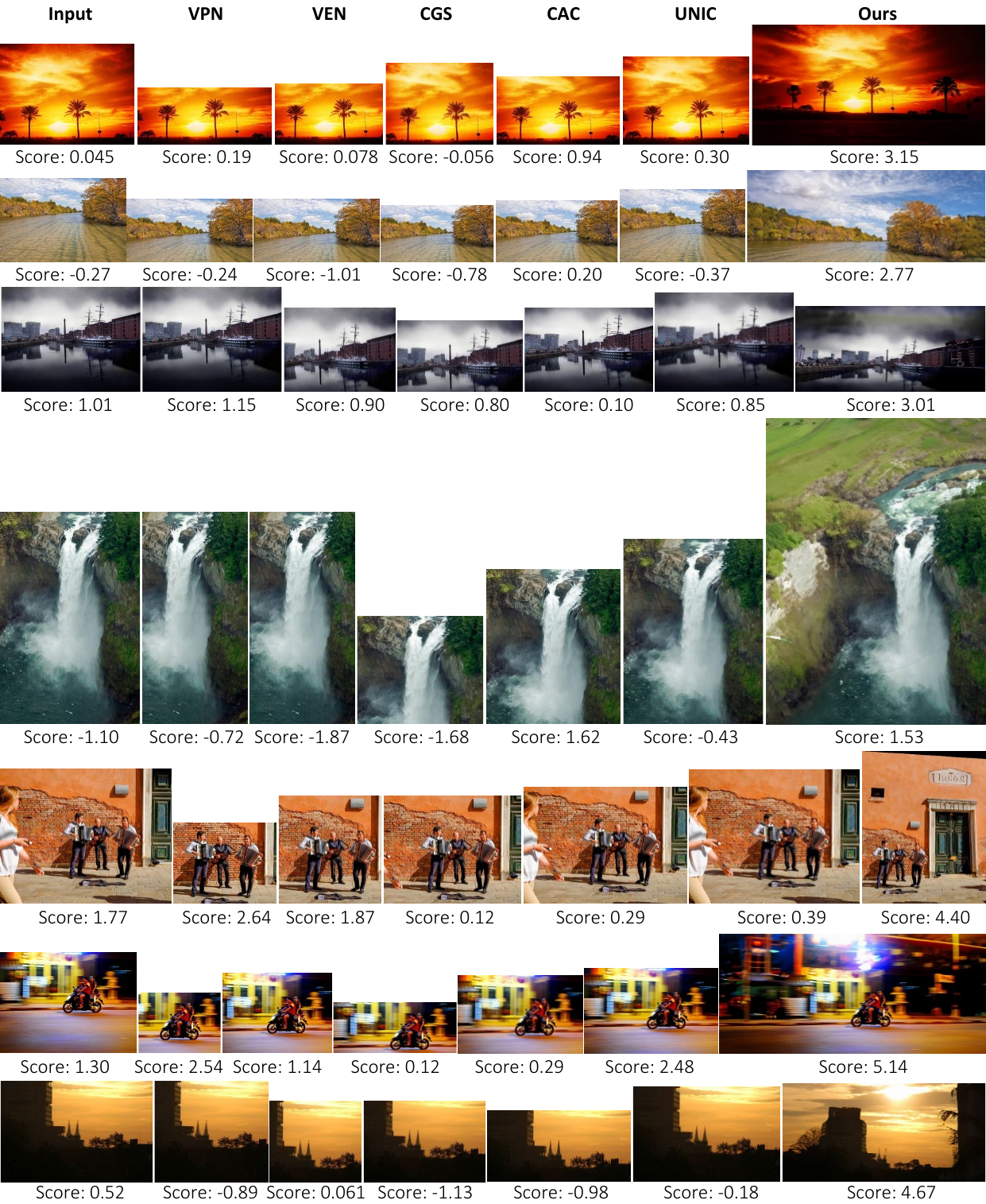}
\caption{Qualitative comparison of our method with existing 2D cropping methods.}
\label{fig:Comparison1}
\end{figure*}

\begin{figure*}[t]
\centering
    \includegraphics[width=\textwidth]{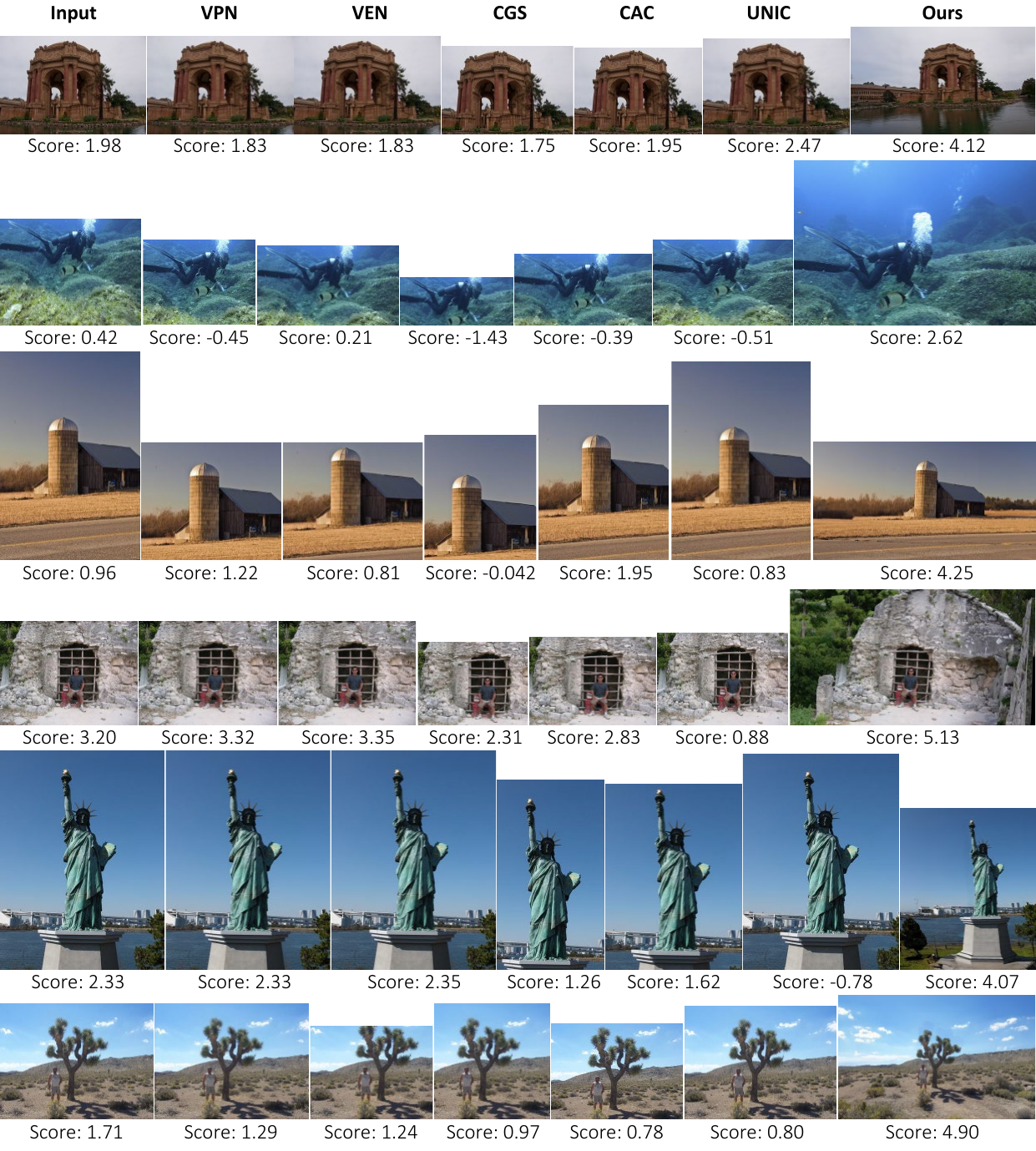}
\caption{Qualitative comparison of our method with existing 2D cropping methods.}
\label{fig:Comparison3}
\end{figure*}

Figures~\ref{fig:Comparison1}, \ref{fig:Comparison2}, and \ref{fig:Comparison3} show the results of qualitative comparison.
The relative image sizes in each row are maintained from the orignal sizes using uniform scaling.
The aesthetic scores under images are calculated using VEN.
It can be observed that the existing methods attempted to extract the most salient objects within the rectangle regions, while our method strived to find views that capture the whole 3D scenes, presumably because VEN prefers such views.
This strategic difference is evident in the decisive differences in the under-image aesthetic scores estimated using VEN. 

\begin{figure*}[!t]
\centering
    \includegraphics[width=\textwidth]{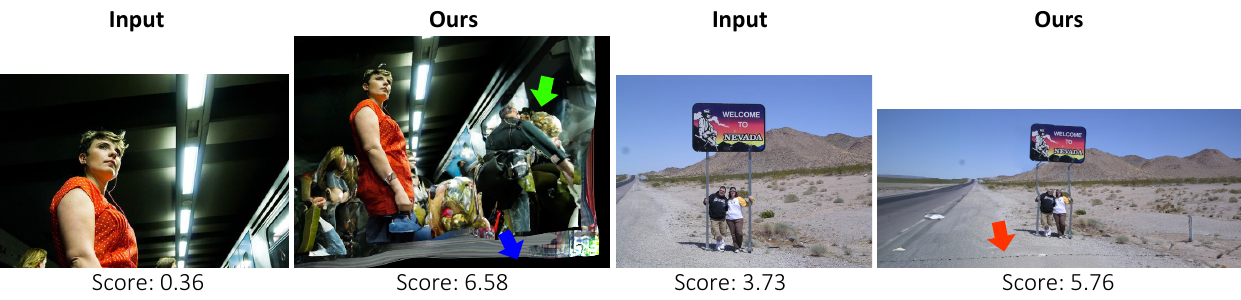}
\caption{Failure cases. Blue arrow: unphotographed regions. Green arrow: unsatisfactory objects created by image extrapolation. Red arrow: gap in the point cloud.}
\label{fig:FailureCases}
\end{figure*}

\subsection{User Study}

To further validate our method, we conducted a user study with 15 test image sets (\ie, seven sets from Figure~\ref{fig:Comparison1}, six from Figure~\ref{fig:Comparison2}, and two from the first to the second rows of Figure~\ref{fig:Comparison3}). 
We compared the top-four methods, including the original images, in reference to Table~\ref{tab:ComparisonScore}, \ie, Input, VEN, UNIC, and Ours.
64 subjects were requested to pick the most aesthetically pleasing result for each test set, mentioning the importance of image composition. 
Consequently, the selection percentages are: 
Input 12.5\%, VEN 24.3\%, UNIC 17.3\%, and Ours 45.9\%, which means that ours outperforms other methods. 

This user study also revealed an interesting tendency of subjects' preferences.
Overall, the subjects preferred images with a wider field of view to grasp the 3D scene structures, while they preferred images focusing on human subjects, if there were any in the scene.
VEN also seemingly prefers a wider field of view, and thus, our optimized images have such views, resulting in the best selection percentage in this user study.
This result stems from the characteristics of VEN, and a comprehensive investigation including other aesthetic score estimators is beyond the scope of our paper, which should be an exciting future work.


\subsection{Limitations}

Currently, our method has the following limitations.

\paragraph{Dependency on image extrapolation quality.} Our method heavily relies on the quality of prior image extrapolation. We currently use Adobe Photoshop's generative fill, but it often yields unsatisfactory extrapolated objects (see the green arrow in Figure~\ref{fig:FailureCases}), particularly when sufficient information for extrapolation is unavailable in the original image. More advanced extrapolation techniques would alleviate this defect in the future.

\paragraph{Appearance of out-of-photographed regions.} Although we introduced a mask regularization term to suppress the appearance of out-of-photographed regions (see Section~\ref{sec:MaskRegularization}), unphotographed regions (\ie the black background) do appear in the rendered images when the camera moves significantly (see the blue arrow in Figure~\ref{fig:FailureCases}). Image extrapolation with much larger regions might sometimes help, but it also increases the risk of the appearance of unsatisfactory objects. A hard constraint defined by the viewing frustum of the initial camera might fix the problem.

\paragraph{Dependency on 3D reconstruction and rendering.} Our method also relies on the quality of 3D reconstruction and its rendering. Distorted objects appear if the quality of 3D reconstruction is low. Also, we observe gaps on object surfaces (see the red arrow in Figure~\ref{fig:FailureCases}) because the point cloud is locally sparse or looks sparse when viewed from specific viewpoints. A view-dependent splatting-based technique might fill such gaps.

\section{Conclusions}

In this paper, we have explored an unprecedented approach in improving image aesthetics by optimizing (i) camera parameters in the 3D scene reconstructed from a single input image and (ii) the output image size. We showed that a gradient-based optimizer easily falls into local minima, while a black-box optimizer successfully finds good solutions. Our resultant images often have drastically different views from those of the input images, which cannot be accomplished using traditional 2D cropping methods. Qualitative and quantitative evaluations as well as the user study demonstrated that our method is more effective in improving image aesthetics than existing methods. 

We are confident that our study is a siginificant first step toward the concept of ``\textit{aesthetic retrospective rephotography},'' which means that we can re-take a photograph by going back to the time of photoshooting to change the camera settings for more aesthetically pleasing photography.
In future work, we would like to explore the vast search space of various camera settings even further, \eg, bokeh with depth-of-field control, tone and contrast control with different exposure levels and camera response curves, motion blur, and vignette effects. 


{
    \small
    \bibliographystyle{ieeenat_fullname}
    \bibliography{main}

\begin{thebibliography}{25}
\providecommand{\natexlab}[1]{#1}
\providecommand{\url}[1]{\texttt{#1}}
\expandafter\ifx\csname urlstyle\endcsname\relax
  \providecommand{\doi}[1]{doi: #1}\else
  \providecommand{\doi}{doi: \begingroup \urlstyle{rm}\Url}\fi

\bibitem[Akiba et~al.(2019)Akiba, Sano, Yanase, Ohta, and Koyama]{optuna_2019}
Takuya Akiba, Shotaro Sano, Toshihiko Yanase, Takeru Ohta, and Masanori Koyama.
\newblock Optuna: {A} next-generation hyperparameter optimization framework.
\newblock In \emph{Proceedings of the 25th {ACM} {SIGKDD} International
  Conference on Knowledge Discovery {\&} Data Mining, {KDD} 2019, Anchorage,
  AK, USA, August 4-8, 2019}, pages 2623--2631. {ACM}, 2019.

\bibitem[AlZayer et~al.(2021)AlZayer, Lin, and Bala]{9636788}
Hadi AlZayer, Hubert Lin, and Kavita Bala.
\newblock Autophoto: Aesthetic photo capture using reinforcement learning.
\newblock In \emph{2021 IEEE/RSJ International Conference on Intelligent Robots
  and Systems (IROS)}, pages 944--951, 2021.

\bibitem[Avidan and Shamir(2007)]{DBLP:journals/tog/AvidanS07}
Shai Avidan and Ariel Shamir.
\newblock Seam carving for content-aware image resizing.
\newblock \emph{{ACM} Trans. Graph.}, 26\penalty0 (3):\penalty0 10, 2007.

\bibitem[Badki et~al.(2017)Badki, Gallo, Kautz, and
  Sen]{DBLP:journals/tog/BadkiGKS17}
Abhishek Badki, Orazio Gallo, Jan Kautz, and Pradeep Sen.
\newblock Computational zoom: a framework for post-capture image composition.
\newblock \emph{{ACM} Trans. Graph.}, 36\penalty0 (4):\penalty0 46:1--46:14,
  2017.

\bibitem[Chen et~al.(2017{\natexlab{a}})Chen, Huang, Chang, Tsai, Chen, and
  Chen]{chen-wacv2017}
Yi{-}Ling Chen, Tzu{-}Wei Huang, Kai{-}Han Chang, Yu{-}Chen Tsai, Hwann{-}Tzong
  Chen, and Bing{-}Yu Chen.
\newblock Quantitative analysis of automatic image cropping algorithms: {A}
  dataset and comparative study.
\newblock In \emph{{WACV} 2017}, pages 226--234. {IEEE} Computer Society,
  2017{\natexlab{a}}.

\bibitem[Chen et~al.(2017{\natexlab{b}})Chen, Klopp, Sun, Chien, and
  Ma]{10.1145/3123266.3123274}
Yi-Ling Chen, Jan Klopp, Min Sun, Shao-Yi Chien, and Kwan-Liu Ma.
\newblock Learning to compose with professional photographs on the web.
\newblock In \emph{Proceedings of the 25th ACM International Conference on
  Multimedia}, page 37–45, New York, NY, USA, 2017{\natexlab{b}}. Association
  for Computing Machinery.

\bibitem[Fang et~al.(2014)Fang, Lin, Mech, and Shen]{10.1145/2647868.2654979}
Chen Fang, Zhe Lin, Radomir Mech, and Xiaohui Shen.
\newblock Automatic image cropping using visual composition, boundary
  simplicity and content preservation models.
\newblock In \emph{Proceedings of the 22nd ACM International Conference on
  Multimedia}, page 1105–1108, New York, NY, USA, 2014. Association for
  Computing Machinery.

\bibitem[Hansen and Ostermeier(2001)]{hansen2001completely}
Nikolaus Hansen and Andreas Ostermeier.
\newblock Completely derandomized self-adaptation in evolution strategies.
\newblock \emph{Evolutionary computation}, 9\penalty0 (2):\penalty0 159--195,
  2001.

\bibitem[He et~al.(2022)He, Zhang, Xie, Jiang, and
  Ming]{DBLP:conf/ijcai/HeZXJM22}
Shuai He, Yongchang Zhang, Rui Xie, Dongxiang Jiang, and Anlong Ming.
\newblock Rethinking image aesthetics assessment: Models, datasets and
  benchmarks.
\newblock In \emph{Proceedings of the Thirty-First International Joint
  Conference on Artificial Intelligence, {IJCAI} 2022, Vienna, Austria, 23-29
  July 2022}, pages 942--948, 2022.

\bibitem[Hong et~al.(2021)Hong, Du, Xian, Lu, Cao, and Zhong]{Hong_2021_CVPR}
Chaoyi Hong, Shuaiyuan Du, Ke Xian, Hao Lu, Zhiguo Cao, and Weicai Zhong.
\newblock Composing photos like a photographer.
\newblock In \emph{Proceedings of the IEEE/CVF Conference on Computer Vision
  and Pattern Recognition (CVPR)}, pages 7057--7066, 2021.

\bibitem[Johnson et~al.(2020)Johnson, Ravi, Reizenstein, Novotny, Tulsiani,
  Lassner, and Branson]{10.1145/3415263.3419160}
Justin Johnson, Nikhila Ravi, Jeremy Reizenstein, David Novotny, Shubham
  Tulsiani, Christoph Lassner, and Steve Branson.
\newblock Accelerating {3D} deep learning with {PyTorch3D}.
\newblock In \emph{SIGGRAPH Asia 2020 Courses}. Association for Computing
  Machinery, 2020.

\bibitem[Kingma and Ba(2015)]{DBLP:journals/corr/KingmaB14}
Diederik~P. Kingma and Jimmy Ba.
\newblock Adam: {A} method for stochastic optimization.
\newblock In \emph{3rd International Conference on Learning Representations,
  {ICLR} 2015, San Diego, CA, USA, May 7-9, 2015, Conference Track
  Proceedings}, 2015.

\bibitem[Li et~al.(2018)Li, Wu, Zhang, and Huang]{li2018a2}
Debang Li, Huikai Wu, Junge Zhang, and Kaiqi Huang.
\newblock A2-rl: Aesthetics aware reinforcement learning for image cropping.
\newblock In \emph{Proceedings of the IEEE conference on computer vision and
  pattern recognition}, pages 8193--8201, 2018.

\bibitem[Li et~al.(2020)Li, Zhang, Huang, and Yang]{li2020composing}
Debang Li, Junge Zhang, Kaiqi Huang, and Ming-Hsuan Yang.
\newblock Composing good shots by exploiting mutual relations.
\newblock In \emph{Proceedings of the IEEE/CVF Conference on Computer Vision
  and Pattern Recognition}, pages 4213--4222, 2020.

\bibitem[Liu et~al.(2010)Liu, Jin, and Wu]{DBLP:conf/cae/LiuJW10}
Ligang Liu, Yong Jin, and Qingbiao Wu.
\newblock Realtime aesthetic image retargeting.
\newblock In \emph{6th International Symposium on Computational Aesthetics in
  Graphics, Visualization, and Imaging (CAe 2010)}, pages 1--8. Eurographics
  Association, 2010.

\bibitem[Liu et~al.(2022)Liu, Agrawala, DiVerdi, and
  Hertzmann]{liu2022zoomshop}
Sean~J Liu, Maneesh Agrawala, Stephen DiVerdi, and Aaron Hertzmann.
\newblock {ZoomShop}: Depth-aware editing of photographic composition.
\newblock In \emph{Computer Graphics Forum}, pages 57--70. Wiley Online
  Library, 2022.

\bibitem[Liu et~al.(2023)Liu, Liu, Li, Liu, Wang, Lei, and Zuo]{Liu_2023_ICCV}
Xiaoyu Liu, Ming Liu, Junyi Li, Shuai Liu, Xiaotao Wang, Lei Lei, and Wangmeng
  Zuo.
\newblock Beyond image borders: Learning feature extrapolation for unbounded
  image composition.
\newblock In \emph{Proceedings of the IEEE/CVF International Conference on
  Computer Vision (ICCV)}, pages 13023--13032, 2023.

\bibitem[Ranftl et~al.(2022)Ranftl, Lasinger, Hafner, Schindler, and
  Koltun]{Ranftl2020}
Ren{\'{e}} Ranftl, Katrin Lasinger, David Hafner, Konrad Schindler, and Vladlen
  Koltun.
\newblock Towards robust monocular depth estimation: Mixing datasets for
  zero-shot cross-dataset transfer.
\newblock \emph{{IEEE} Trans. Pattern Anal. Mach. Intell.}, 44\penalty0
  (3):\penalty0 1623--1637, 2022.

\bibitem[Rubinstein et~al.(2010)Rubinstein, Gutierrez, Sorkine, and
  Shamir]{10.1145/1882261.1866186}
Michael Rubinstein, Diego Gutierrez, Olga Sorkine, and Ariel Shamir.
\newblock A comparative study of image retargeting.
\newblock \emph{ACM Trans. Graph.}, 29\penalty0 (6), 2010.

\bibitem[Shih et~al.(2020)Shih, Su, Kopf, and Huang]{Shih3DP20}
Meng{-}Li Shih, Shih{-}Yang Su, Johannes Kopf, and Jia{-}Bin Huang.
\newblock {3D} photography using context-aware layered depth inpainting.
\newblock In \emph{{CVPR} 2020}, pages 8025--8035. Computer Vision Foundation /
  {IEEE}, 2020.

\bibitem[Wei et~al.(2018)Wei, Zhang, Shen, Lin, Mech, Hoai, and
  Samaras]{wei-cvpr2018}
Zijun Wei, Jianming Zhang, Xiaohui Shen, Zhe Lin, Radomír Mech, Minh Hoai, and
  Dimitris Samaras.
\newblock Good view hunting: Learning photo composition from dense view pairs.
\newblock In \emph{2018 IEEE/CVF Conference on Computer Vision and Pattern
  Recognition}, pages 5437--5446, 2018.

\bibitem[Xie et~al.(2023)Xie, Hu, Sun, Pirk, Zhang, Mech, and
  Kaufman]{xie2023gait}
Desai Xie, Ping Hu, Xin Sun, Soren Pirk, Jianming Zhang, Radom{\'\i}r Mech, and
  Arie~E Kaufman.
\newblock Gait: Generating aesthetic indoor tours with deep reinforcement
  learning.
\newblock In \emph{Proceedings of the IEEE/CVF International Conference on
  Computer Vision}, pages 7409--7419, 2023.

\bibitem[Zeng et~al.(2019)Zeng, Li, Cao, and Zhang]{8953733}
H. Zeng, L. Li, Z. Cao, and L. Zhang.
\newblock Reliable and efficient image cropping: A grid anchor based approach.
\newblock In \emph{2019 IEEE/CVF Conference on Computer Vision and Pattern
  Recognition (CVPR)}, pages 5942--5950, Los Alamitos, CA, USA, 2019. IEEE
  Computer Society.

\bibitem[Zhang et~al.(2021)Zhang, Niu, and Zhang]{zhang2021image}
Bo Zhang, Li Niu, and Liqing Zhang.
\newblock Image composition assessment with saliency-augmented multi-pattern
  pooling.
\newblock In \emph{{BMVC} 2021}, page 144. {BMVA} Press, 2021.

\bibitem[Zhong et~al.(2021)Zhong, Li, Huang, Zhang, Lu, and
  Wang]{zhong2021aesthetic}
Lei Zhong, Feng-Heng Li, Hao-Zhi Huang, Yong Zhang, Shao-Ping Lu, and Jue Wang.
\newblock Aesthetic-guided outward image cropping.
\newblock \emph{ACM Transactions on Graphics (TOG)}, 40\penalty0 (6):\penalty0
  1--13, 2021.

\end{thebibliography}
}

\end{document}